# Improving Word Vector with Prior Knowledge in Semantic Dictionary

Wei Li[1]   Yunfang Wu[1(✉)]   Xueqiang Lv[2]

[1] Key Laboratory of Computational Linguistics, Peking University, Beijing
`liweitj47@pku.edu.cn`  `wuyf@pku.edu.cn`
[2] Beijing Key Laboratory of Internet Culture and Digital Dissemination Research, Beijing
`lxq@bistu.edu.cn`

**Abstract.** Using low dimensional vector space to represent words has been very effective in many NLP tasks. However, it doesn't work well when faced with the problem of rare and unseen words. In this paper, we propose to leverage the knowledge in semantic dictionary in combination with some morphological information to build an enhanced vector space. We get an improvement of 2.3% over the state-of-the-art Heidel Time system in temporal expression recognition, and obtain a large gain in other name entity recognition (NER) tasks. The semantic dictionary Hownet alone also shows promising results in computing lexical similarity.

## 1 Introduction

To get a good representation of words has been a fundamental task for many natural language processing (NLP) tasks. The bag of words model (BOW) has been a practical and handy way. However, it cannot encode any information within the representation itself. With the help of neural networks, researchers are now able to represent words with distributed low dimensional vectors. Mikolov et al. (2013) proposed continuous bag of words model (CBOW) and continuous skip gram model to learn the word vectors, which have been very successful. However, it still suffers from the sparsity problem, because this kind of model learns word embeddings entirely on the basis of contexts, leading to unreliable representations for rare words.

To deal with the sparsity problem, Luong et al. (2013) and Qiu et al. (2014) proposed to consider the words and their morphemes together when building the vector space. Cui et al. (2015) tried to seek morphological information to revise the vectors of rare words in English. Chen et al. (2015) proposed Character-enhanced word embedding model (CWE) to combine words and multiple prototype character embedding together. Sun et al. (2014) and Li et al. (2015) proposed to leverage the component (or radical) information to learn Chinese character embedding. Peng and Dredze (2015) proposed a joint training objective for NER tasks in social media corpus with character embedding. These kinds of models all try to introduce the information of inner structure of a word to supplement the word level embedding.

In this paper, we propose to apply the sememes in Hownet (Dong and Dong, 2006) together with morphological information to deal with the sparsity problem. In our opinion, sematic dictionaries like Hownet are very valuable resources. To utilize the knowledge of the dictionary, we calculate the embedding of each sememe whose target words appear in training corpus and then for each word, we sum the sememes of the word according to the semantic dictionary to form an approximate embedding of the word. Experiments show that this produces a highly correlated description with human judgment. As far as we know, this is the first time that the knowledge from a semantic dictionary like Hownet sememes been introduced to word vector space.

We apply our new word embeddings to tackle the problem of recognizing temporal expressions, person names and locations. Compared with the state-of-the-art F value of 87.3 from Heidel Time (Li et al., 2014), We got an F value of 89.6 on temeval-2 data concerning temporal expression recognition, improving the F value by 2.3 without any hand crafted feature. We also achieve competitive results on recognizing person names and locations.

## 2   Our approach

Our approach is shown in **Fig. 1**. First, we get the *original word vector space* produced by word2vec. Then, we use the morphological information to find the similar words for a target word, the rare and unseen words. After that, we get the *similar word vector space* based on the word embeddings of the similar words. Then we update the vector space by combining the original vector and the similar words' vectors with weights tuned by term frequency. We concatenate this new vector with a context window of 2, along with *Hownet vector* and *Character vector* which we will come to later, and get the final *complete vector space*. The complete vectors are then used as real-number features to predict the tag of the word with multi-class logistic regression.

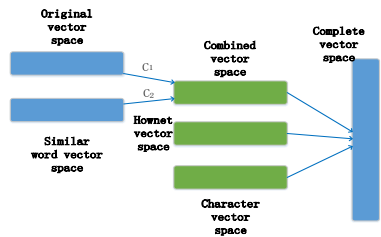

**Fig. 1.** Word embedding for rare words

### 2.1   Hownet Semantic Knowledge

Hownet is an influential semantic dictionary in Chinese. It describes the meaning of a word by a series of concepts. The concepts are further explained by a finite set of 1,500 sememes from ten classes and the relationship between them. An example is shown in **Table 1**.

Table 1. Example of Hownet description

| Chinese | POS | English | Description |
|---|---|---|---|
| 房租 | N | House rent | expenditure \|费用,*borrow\|借入, # house \|房屋 |

In this example, "房租" (House rent) is the target word, "费用(expenditure),借入(borrow),房屋(house)" are the corresponding sememes. The symbols "*,#" indicate the relations between the sememes, which we omit for now. On one hand this type of representation loses some details and can only give a rough description, on the other hand we believe it brings generality to the space.

1. Ideally, we want to use existing embeddings of sememes trained from real corpus. However, many of the sememes are rare or unseen. For instance, the second sememe in our example, "借入". Though its meaning is very clear, it can be rarely seen in real life corpus as it is too formal. So we need to get a good sememe embedding first.
2. To learn sensible sememe embeddings, we replace the words with their first sememe into the training corpus, and train the embedding using word2vec. For instance in our example, all occurrences of "房租" in the corpus would all be replaced by "费用", and thus we can get the embedding of the sememe "费用", $\theta("费用")$. In the dictionary, the sememe "费用" appears not only in "房租", but also in a lot of other words like "安家费" (settlement allowance), so $\theta("费用") \neq \theta("房租")$. The first sememes are all basic sememes, containing most of the meaning.
3. We do the same thing for the second and the third sememes of each word. For instance, "借入","房屋" will be put back to the corpus and replace the original word "房租" separately and get their according embeddings $\theta("借入")$ and $\theta("房屋")$. Because these sememes appear in different words, they would have different word embeddings, that is to say, $\theta("费用")$, $\theta("借入")$ and $\theta("房屋")$ are different. By getting the Hownet embeddings this way, we only hold the basic sememes of each word. This trick is handy and pragmatic, but it also makes the Hownet embedding lose some detail, which can cause ambiguity.
4. Inspired by the attribute of word vectors observed in Mikolov et al. (2013) that simple algebraic addition can lead to very interesting results, for example, $\theta("Germany")+\theta("capital") \approx \theta("Berlin")$. We sum the embeddings of all sememes of a word to get a new embedding, and we call it the Hownet embedding. Concretely, in our example, we use $\theta("费用") + \theta("借入") + \theta("房屋")$ as an approximation to $\theta("房租")$.

With the method above, we can produce word embeddings out of the sememe embeddings and we call this the *Hownet vector space* in **Fig. 1**.

## 2.2 Similar Words

In Chinese, the vocabulary size of characters is much smaller than that of words, which means it is much rarer to meet new characters than new words. Apart from that, most of the words consist of only two to three characters, especially in nouns. Based on these observations and the assumption that morphologically similar words tend to be semantically similar, we want to find the morphologically similar words of rare and unseen words to help learn or revise their embeddings.

We applied longest common substring (LCS), edit distance, cosine similarity to measure the morphological similarity between words. After we get the similarity scores from these three measurements, we use a simple perceptron to tune their weights.

1. We use another semantic dictionary in Chinese called "*tong yi ci ci lin*" to build the training set. This dictionary contains the categorical information about the synonymous sets of words. The pairs of words in the training set are selected either from the same category or randomly from different categories. If two words belong to the same category, which means they are synonyms, we set the similarity to be *1*. If they belong to different categories, we set the similarity score to be *0*.
2. We use the method aforementioned to find the similarity scores between the rare or unseen words in the new coming corpus and the words in the existing vocabulary (words from the training set). We put the top five words that have the highest similarity scores into the candidate set.
3. We use term frequency of words in the training set as weights to calculate a weighted average of these candidate word embeddings. We then set this new embedding to be the embedding of the rare or unseen words.

As the frequency of some words can be very large, giving them too much significance, we use a function *f* consists of five buckets to separate these words with different weights rather than use their term frequency (*tf*) directly as is shown below.

$$f(tf) = \begin{cases} 4 & \text{if } tf > 100 \\ 3 & \text{if } tf > 20 \text{ and } tf <= 100 \\ 2 & \text{if } tf > 5 \text{ and } tf <= 20 \\ 1 & \text{if } tf > 2 \text{ and } tf <= 5 \\ 0 & \text{if } tf <= 2 \end{cases}$$

After the revision, we get a new version of the embedding space of rare and unseen words which is the *Similar word vector space* in **Fig. 1**. We can then combine this space with the *original vector space* produced by word2vec as a complement to form a new *combined vector space* (**Fig. 1**). When we combine these two vector spaces, we also use two weights $C_1$ and $C_2$ in **Fig. 1** to weighted sum the two embeddings. The weights here also depend on term frequency.

### 2.3 Character embedding

Besides the fact that similar words should have similar meanings and similar embeddings, the inner structure of a word also provides valuable information. It is observed that the last character of a word holds most part of the meaning for most of nouns. This characteristic is very useful in the problem of recognizing temporal expressions and name entity words. For instance, in the temporal expression "4月5日", "日" is a very clear sign indicating that the word should be a date. To get the character level embedding, we feed the separate characters instead of the original words in the same training corpus to word2vec. We call this space of character level embedding the *character vector space*.

## 3 Experiment

### 3.1 Correlation with human judgement

Word embedding has been proven to be successful in measuring the similarity or relatedness between words. In this experiment, we show that the cosine similarity calculated with the Hownet embedding show a good correlation with human judgement. We design a questionnaire of forty pairs of words. The words are all selected from Hownet, and at least one of the words in each pair is not observed in our training data. We use the corpus of three-month People Daily to be the training data.

In the experiment, we ask ten students to mark the similarity between words, ranging from 1 to 10. We average the ten scores, and set the mean as the human judgement. Then we calculate the cosine similarity within the word pairs based on *Hownet embedding* we get, and use **Spearman** to measure the correlation between Hownet embedding similarity and human judgement, obtaining a Spearman of 56.6. We also implement the similarity computing method of Liu and Li (2002) (also based on Hownet) with all default settings in the same data, and get a Spearman of only 38.0. Some examples are shown in Table 2, where Word 1 is the rare word.

This result shows that the embedding learnt from Hownet has a fairly strong correlation with human judgement. While most of the cosine similarity between our Hownet embedding and human judgement result correlate with each other, some of the words failed. We think this is partly due to the fact that we dropped the last sememes of the words, and this way of representing words loses the details of the words.

Table 2. Examples of Hownet experiment

| Word 1 | Word 2 | Human | Hownet | Liu & Li (2002) |
|---|---|---|---|---|
| 殃及 | 到达 | 2.3 | 0.021 | 0.138 |
| 伤残人 | 敌人 | 2.6 | 0.02 | 0.722 |
| 六月份 | 时间 | 5.6 | 0.451 | 0.6 |
| 薪水 | 工资 | 9.2 | 1 | 1 |
| 活该 | 应该 | 3.2 | 0.332 | 0.074 |
| 次序 | 秩序 | 4.7 | 1 | 1 |
| 衣料 | 材料 | 5.3 | 0.143 | 1 |
| 阿拉伯人 | 阿拉伯 | 7.2 | 0.106 | 0.149 |
| 妥当 | 合适 | 7.8 | 0.168 | 1 |
| 找出 | 发现 | 7.4 | 0.268 | 0.55 |

One point should be stated is that although Hownet contains only a limited number of words, our main purpose is not using any specific dictionary, but to introduce this new way of constructing embeddings or to bring the information into the vector space.

### 3.2 Temporal Tagging

We focus on the recognition part of Chinese temporal expressions in Tempeval-2 (Verhagen et al., 2010). This part of the data consists of 931 sentences in the train set, 345 sentences in the test set, which were all newspaper texts and correctly segmented. There are four tags, namely, Time, Date, Duration, Set, following the timex3 standard. There are 936 temporal expressions in total. In our experiment, we tag the words with "BI" system. Because of the shortage of data, the distribution of the tags in the data is highly skewed and can be seen in **Fig. 2**.

For a long time, researchers have been in favor of rule-based methods to do the job, and got a fairly good result. Li et al. (2014) created a rule-based Heidel Time system adapted from English and got the art-of-the-art F value of 87.3. We train the embedding with word2vec on the corpus of three-month People Daily News together with the train data provided in Temevel-2, revise it to get an enhanced embedding with the method aforementioned.

Considering the fact that the training data in temeval-2 is rather small, we feed the vectors into a log-linear classifier (Fan et al., 2008) with L2 regularization to do the work to avoid overfitting. We deal with the problem as a multi-class classification (logistic regression) problem, and the vectors learnt from our models can be put forward to the log-linear classifier directly as real number features. We choose a context window of 2 concatenated with our modified vector.

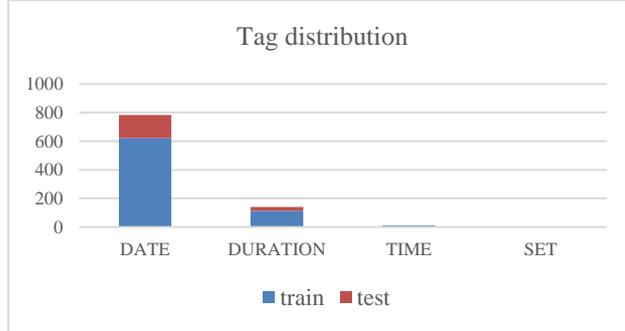

**Fig. 2.** Tag distribution in Tempeval2 task

The settings of our baselines are listed below.

- **Word2vec**: This method feeds the embedding learnt with word2vec (*original vector space*) with a context window of 2 directly to the log-linear classifier.
- **Character**: This method concatenates the embedding from the *original vector space* in a window of 2 and the embedding of the last character of the word (*character vector space*), then feeds this new embedding to the classifier.
- **Morphological**: This method feeds the embedding from *combined vector space* with a context window of 2 to the classifier.
- **Hownet**: This method concatenates the embedding from the *original vector space* in a window of 2 with the embedding from *Hownet vector space*, and feeds this new embedding to the classifier.
- **Final model**: This method feeds the embedding from the *complete vector space*, which is the concatenation of embeddings from *combined vector space*, *character vector space* and *Hownet vector space*, to the classifier.

**Table 3.** Temp2eval results

| Method | P | R | F |
|---|---|---|---|
| Heidel Time | 93.4 | 82 | 87.3 |
| Word2vec | 86.5 | 74.1 | 79.8 |
| Character | 85.0 | 81.4 | 83.2 |
| Morphological | 86.6 | 74.7 | 80.2 |
| Hownet | 87.0 | 79.6 | 83.1 |
| Final Model | 93.7 | 86.0 | 89.6 |

The results are shown in Table 3. It shows that our proposed Hownet embedding brings a big improvement compared to the original word2vec space in F value (**79.8** to **83.1**). And with the help of other methods, the overall result beats the Heidel Time baseline. It can also be seen that the overall method gives a big improvement to the original word2vec method, which indicates that the revised vector space captures the information better than the original one. Still, the lack of tagged data and the skewness of the tags strongly hold the performance back.

### 3.3 People Daily NER

Person names and locations have been a central part in name entity recognition problem. In this paper, we use an open corpus of 10 days of People Daily (1998) as our train and test data, and test data is extracted with five-fold fashion. This data was segmented and tagged with part of speech information by the institute of Computational Linguistics of Peking University, and can be freely downloaded. We build the test set using the POS tags of "ns" and "nr", and so when training the classifier, POS tags are omitted. We predict ns and nr tags with our enhanced embedding, based only on the segmented words. The results are shown in **Table 4**. We also treat this NER problem as a multi-class logistic regression problem with the vectors as real-number features. The settings of the baselines are the same as in tempeval2 task.

In both these two tasks, our enhanced embedding performs much better than the original version of word2vec. And Hownet embedding gives big improvement over the original word2vec vector space by an F score of almost *5* percent. We think this improvement comes not only from the Hownet and character embedding for the rare and unseen words, but also from the generality these two embeddings bring to the original embeddings. This phenomenon is true especially when one cannot get a big enough training data.

Table 4. NER results on People Daily News

| Method | NR | | | NS | | |
|---|---|---|---|---|---|---|
| | P | R | F | P | R | F |
| Word2vec | 92.6 | 83.4 | 87.8 | 82.6 | 65.3 | 72.9 |
| Character | 92.0 | 86.1 | 88.9 | 83.6 | 73.3 | 78.1 |
| Similar words | 90.9 | 85.8 | 88.3 | 80.4 | 67.3 | 73.3 |
| Hownet | 93.4 | 86.3 | 89.8 | 85.1 | 71.7 | 77.8 |
| Final model | 93.5 | 89.6 | 91.5 | 85.8 | 78.9 | 82.2 |

## 4 Conclusion

Although the deep neural network alone seems very promising in capturing the semantics of words, we believe that the prior knowledge that experts have been working on for decades should not be ignored. In this paper, we propose a method of introducing the knowledge of semantic dictionary Hownet together with some morphological information into the vector space. Experiments show that this method is very effective in capturing semantic information, especially when we don't have the access to big training data. As should be pointed out, our way of using semantic dictionary Hownet aims to provide a new perspective of utilizing semantic dictionary information, which should not be limited to this particular dictionary. In the future, we hope to find more efficient and suitable ways to automatically construct these sememe embeddings, and combine them with more sophisticated neural networks.

**Acknowledgement**
This work is supported by National Natural Science Foundation of China (61371129),